\documentclass[twocolumn, 8pt]{article}

\usepackage{lipsum}
\usepackage{JAMIA}
\usepackage{pifont}

\usepackage{tabularx}
\usepackage{longtable}
\usepackage{array}
\usepackage{ragged2e}

\usepackage{amsmath}
\usepackage{amssymb}
\usepackage{mathtools}
\usepackage{amsthm}
\usepackage{url}

\newcommand{\cmark}{\ding{51}}%
\newcommand{\xmark}{\ding{55}}%
\title{Early GVHD Prediction in Liver Transplantation via Multi-Modal Deep Learning on Imbalanced EHR Data}
\author[1,3]{Yushan Jiang, MS}
\author[1]{Shuteng Niu, PhD}
\author[3]{Dongjin Song, PhD}
\author[2]{Yichen Wang, MB, MS}
\author[1]{Jingna Feng, MS}
\author[1]{Xinyue Hu, MS}
\author[2]{Liu Yang, MBBS}
\author[1]{Cui Tao, PhD}

\affil[1]{Department of Artificial Intelligence and Informatics, Mayo Clinic, Jacksonville, FL 32224, United States}
\affil[2]{Gastroenterology and Hepatology Transplant Center, Mayo Clinic, Jacksonville, FL 32224, United States}
\affil[3]{School of Computing, University of Connecticut, Storrs, CT 06269, United States}

\date{}

\begin{document}

\par\noindent\rule[-7pt]{15.5cm}{0.2em}
\begin{strip}
    \begin{minipage}{.88\textwidth}
        \maketitle
        \small
        \abstractSection
        {Graft-versus-host disease (GVHD) is a rare but often fatal complication in liver transplantation, with a very high mortality rate. By harnessing multi-modal deep learning methods to integrate heterogeneous and imbalanced electronic health records (EHR), we aim to advance early prediction of GVHD, paving the way for timely intervention and improved patient outcomes.} 
        { In this study, we analyzed pre-transplant electronic health records (EHR) spanning the period before surgery for 2,100 liver transplantation patients, including 42 cases of graft-versus-host disease (GVHD), from a cohort treated at Mayo Clinic between 1992 and 2025. The dataset comprised four major modalities: patient demographics, laboratory tests, diagnoses, and medications. We developed a multi-modal deep learning framework that dynamically fuses these modalities, handles irregular records with missing values, and addresses extreme class imbalance through AUC-based optimization.} 
        {The developed framework outperforms all single-modal and multi-modal machine learning baselines, achieving an AUC of 0.836, an AUPRC of 0.157, a recall of 0.768, and a specificity of 0.803. It also demonstrates the effectiveness of our approach in capturing complementary information from different modalities, leading to improved performance.} 
        {Our multi-modal deep learning framework substantially improves existing approaches for early GVHD prediction. By effectively addressing the challenges of heterogeneity and extreme class imbalance in real-world EHR, it achieves accurate early prediction.} 
        {Our proposed multi-modal deep learning method demonstrates promising results for early prediction of a GVHD in liver transplantation, despite the challenge of extremely imbalanced EHR data.} 
        {multi-modal model, deep learning, GVHD, early risk prediction, electronic health records} 
        
        \par\noindent\rule[-7pt]{15.5cm}{0.2em}
        \hspace{2cm}
    \end{minipage}
\end{strip}


    



\section*{Introduction}

Graft-versus-host disease (GVHD) in liver transplantation (LT) is a rare yet fatal immune complication, with a reported incidence of approximately 0.5--2\% and mortality rates exceeding 75\% \cite{guo2008graft,wood2020graft,akbulut2012graft}. It arises when donor-derived immune cells mount an attack against recipient tissues, triggering systemic inflammation and multi-organ dysfunction. Early detection of GVHD in liver transplantation remains particularly challenging. Its extreme rarity limits large-scale studies and systematic evaluation of risk factors. In addition, patients often present with nonspecific symptoms such as rash and diarrhea, which are easily mis-attributed to infections, drug toxicity, or acute rejection~\cite{murali2016graft}. The absence of reliable clinical biomarkers or predictive models for pre-transplant risk assessment further hinders early diagnosis. Consequently, GVHD is often recognized only after it has progressed to advanced stages, and such delayed diagnosis contributes to the persistently high mortality, underscoring the urgent need for accurate and timely prediction methods.

Recent advances in electronic health records (EHRs) provide a promising avenue to address this gap. Liver transplant candidates undergo extensive preoperative evaluation, generating rich multi-modal data encompassing demographics, laboratory tests, diagnosis, and medication histories. The collected records enable data-driven clinical discovery beyond traditional clinical assessment~\cite{wang2024recent}. In particular, deep learning methods have demonstrated strong capabilities across diverse clinical prediction tasks on existing EHR benchmarks~\cite{johnson2016mimic,johnson2020mimic,pollard2018eicu}, including mortality prediction~\cite{ma2020concare}, readmission prediction~\cite{wang2023hierarchical}, adverse event detection~\cite{tang2020democratizing,choi2016retain}, and treatment recommendation~\cite{wang2020adversarial,jiang2023interpretable,jiang2023fedskill}. By uncovering complex temporal patterns in patient trajectories, deep learning establishes a foundation for predictive models that may enable earlier identification of GVHD, which has been demonstrated in hematopoietic stem cell transplantations~\cite{tang2020predicting,jo2023convolutional,song2025development}.

However, real-world EHRs are highly heterogeneous, irregularly sampled, and plagued by extensive missingness. In addition, the extreme class imbalance of GVHD outcomes in liver transplant patients makes it particularly difficult to learn robust and discriminative patterns for prediction. These challenges hinder most prior methods developed on relatively organized benchmark datasets, leaving early GVHD prediction in liver transplantation largely underexplored within clinical informatics. To the best of our knowledge, the only related work has focused on post-transplant settings and relied on specialized and donor-recipient information~\cite{cooper2022acute}, which is not readily available. To date, no study has investigated pre-transplant EHR for early GVHD prediction.


To this end, we propose a multi-modal deep learning framework for early GVHD prediction in liver transplantation based solely on pre-transplant EHR data. Our study uses a cohort of 2,100 liver transplant recipients treated at Mayo Clinic between 1992 and 2025, among whom 42 developed GVHD. We integrate four major modalities (\textit{i.e.}, demographics, laboratory tests, diagnoses, and medications) into a unified predictive model. Our main contributions are summarized as follows:
\begin{itemize}
    \item \textbf{Multi-modal framework:} We present the first pre-transplant GVHD prediction framework in liver transplantation, capturing predictive signals across heterogeneous modalities.  
    
    \item \textbf{Robust modeling:} We design modality-specific encoders to handle irregular intervals and substantial missingness, apply cross-modal fusion to extract complementary predictive signals, and incorporate AUC-based optimization to ensure robust learning under extreme class imbalance.  
    
    \item \textbf{Comprehensive validation:} On the Mayo Clinic cohort, our method achieves promising predictive performance (AUC = 0.836, AUPRC = 0.157, recall = 0.768, specificity = 0.803), consistently outperforming both traditional machine learning and state-of-the-art deep learning baselines. Ablation studies further validate the effectiveness of our proposed components, highlighting the benefits of handling irregular intervals and missingness as well as leveraging complementary information across modalities.  
\end{itemize}

Collectively, these results demonstrate that multi-modal heterogeneous EHR modeling with imbalance-aware optimization enables earlier GVHD identification, supporting timely intervention and ultimately improving patient outcomes in liver transplantation.

\section*{Materials and Methods}
\subsection*{Primary Cohort for Multi-modal GVHD Prediction}

We curated a cohort of liver transplantation (LT) recipients treated at the Mayo Clinic between 1992 and 2025 to enable multi-modal prediction of GVHD \cite{abdelhameed2023deep}. All patients who underwent LT during this period were initially eligible. To ensure that all utilized data reflected adult clinical profiles, we excluded patients who were younger than 18 years at any point within the three-year pre-transplant observation window. This criterion is equivalent to excluding individuals younger than 21 years at the time of transplantation. After applying the age filter, the cohort comprised 42 GVHD cases and 8,350 controls. To balance computational feasibility while preserving all positive cases, we retained all 42 GVHD cases and randomly selected 2,058 controls, yielding a final analytic cohort of 2,100 adult LT recipients, of whom 42 (2.0\%) developed GVHD post-transplant. GVHD outcomes were identified from structured diagnostic codes and subsequently validated through manual chart review by transplant hepatologists to ensure clinical accuracy.
The primary prediction target was the occurrence of GVHD following LT, formulated as a binary classification task characterized by severe class imbalance.

All available pre-transplant records were extracted and processed into structured formats as model inputs for each patient. Four major data modalities were used: demographic information, laboratory test results, diagnosis data, and medication administration records. 

\textbf{Demographics.} This modality includes 4 features: gender, race, ethnicity, and age. The median age at transplantation was 59 years (range: 22--76), with a mean of 56.7 years. Overall, 65.1\% of patients were male and 34.9\% female. Most patients identified as White (87.2\%), followed by other race (4.4\%), Asian (3.6\%), Black (3.5\%), and unknown (1.4\%). In terms of ethnicity, 84.0\% were non-Hispanic, 14.0\% Hispanic, and 1.9\% were unknown or declined to answer.

\textbf{Diagnosis.} This modality includes 20 symptom-related diagnoses extracted from structured records up to the three years preceding transplantation. Diagnoses were aggregated every three months as a union of visits, yielding a temporal sequence of diagnosis representations. The included diagnoses span hepatic, respiratory, metabolic, infectious, and treatment-related conditions relevant to transplant risk.

\textbf{Drug.} This modality captures the daily administration of 280 commonly used inpatient drugs during the 24 observations preceding liver transplantation, aggregated at a 1-day interval. For each day, the feature represents the count of administered doses for each drug. The drugs span a wide range of classes, including antibiotics and antimicrobials (\textit{e.g.}, meropenem, vancomycin), electrolyte management agents (\textit{e.g.}, sodium bicarbonate, potassium chloride), analgesics and sedatives (\textit{e.g.}, fentanyl, morphine, midazolam), endocrine/metabolic therapies (\textit{e.g.}, various forms of insulin), immunosuppressants and corticosteroids (\textit{e.g.}, tacrolimus, prednisone), and liver-specific agents (\textit{e.g.}, lactulose, rifaximin). The remaining drugs encompass a heterogeneous group of pharmacologic categories.

\textbf{Lab Test.} This modality includes 74 laboratory test variables, extracted using a fixed 1-hour context window across the 24 most recent observations prior to transplantation. Each observation captures the last available test result within the corresponding window. The selected tests span diverse clinical categories, including liver function (\textit{e.g.}, AST, ALT, bilirubin), renal function (\textit{e.g.}, BUN, creatinine), coagulation (\textit{e.g.}, INR, APTT, fibrinogen), hematology (\textit{e.g.}, hemoglobin, platelets, WBC), and electrolytes/metabolites (\textit{e.g.}, sodium, potassium, bicarbonate, calcium). The overall missing ratio was 72.9\%, reflecting the inherent sparsity of laboratory data due to on-demand and irregular testing patterns in real-world clinical practice. Note that all three temporal modalities were constructed from records spanning one or multiple clinical visits. As such, the time steps are not necessarily consecutive and may reflect irregular intervals between observations.

The study protocol was approved by the Mayo Clinic Institutional Review Board (IRB), and all data access and analyses were conducted in accordance with institutional guidelines, HIPAA regulations, and applicable ethical standards.
\begin{table}[ht]
\centering
\setlength{\tabcolsep}{4.2pt}
\caption{Summary statistics across all modalities.}
\begin{tabular}{lcccc}
\hline
\textbf{Modality}     & \textbf{\#Features} & \textbf{\#Steps} & \textbf{Interval} \\
\hline
Demographics & 4   & -- & --     \\
Lab Test     & 74  & 24 & 1-hour  \\
Diagnosis    & 20  & 12 & 3-month \\
Drug         & 280 & 24 & 1-day   \\
\hline
\end{tabular}
\label{tab::data_summary}
\end{table}

\subsection*{Multi-modal Deep Learning Framework for GVHD Prediction in Liver Transplantation}

\begin{figure*}
    \centering
    \includegraphics[width=1\linewidth]{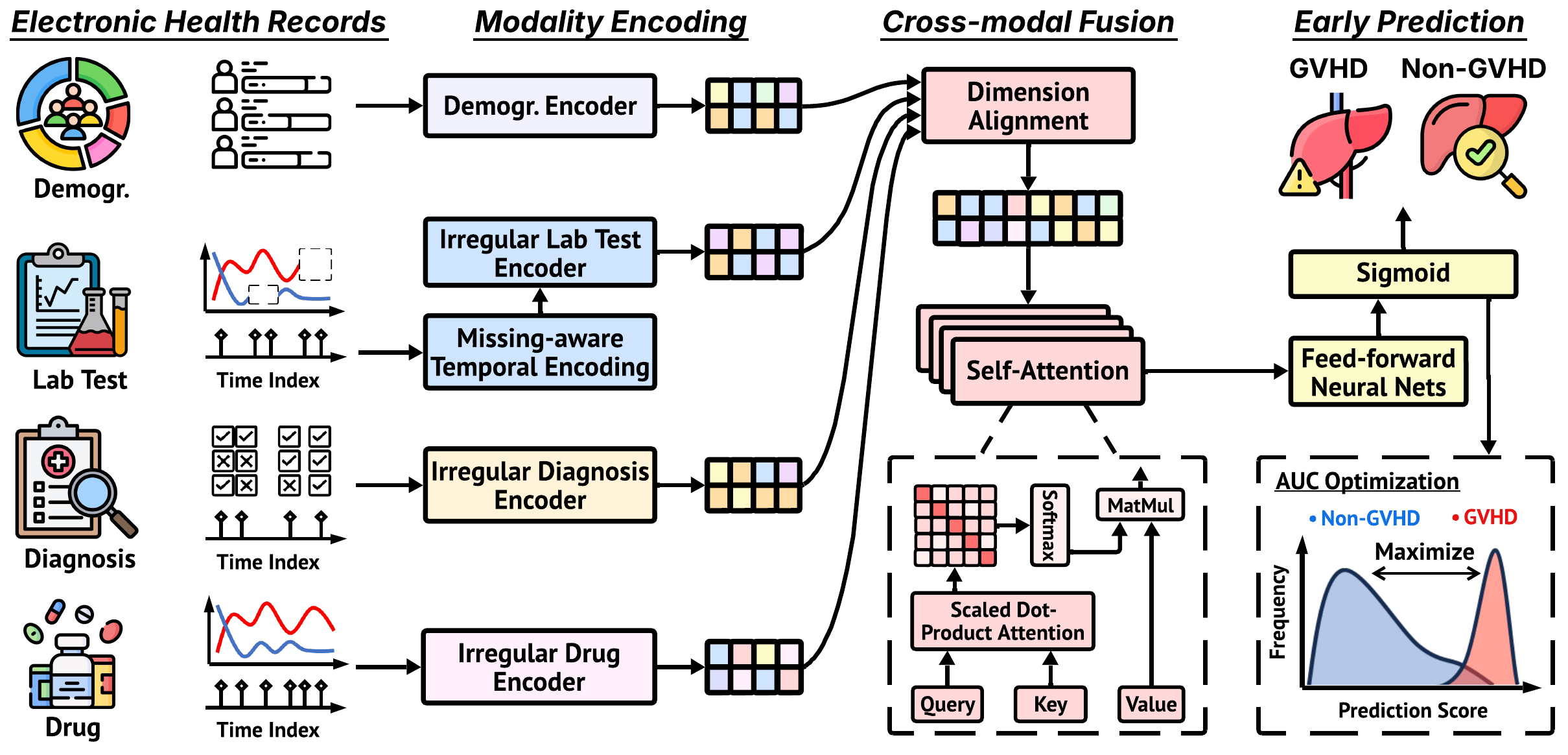}
    \caption{Early prediction of Graft-versus-host disease (GVHD) in liver transplantation using multi‑modal EHRs.}
    \label{fig:framework}
\end{figure*}

\noindent \textbf{Notation.} First, we introduce the notations for the model inputs of each modality. We denote the whole EHR input as:
$
\mathbf{X} = \big\{\mathbf{X}^{\mathrm{demo}},\, \mathbf{X}^{\mathrm{lab}},\, \mathbf{X}^{\mathrm{dx}},\, \mathbf{X}^{\mathrm{drug}}\big\}
$, 
where $\mathbf{X}^{\mathrm{demo}} \in \mathbb{R}^{F_{\mathrm{de}}}$, $\mathbf{X}^{\mathrm{lab}} \in \mathbb{R}^{T_{\mathrm{lab}} \times F_{\mathrm{lab}}}$, $\mathbf{X}^{\mathrm{dx}} \in \mathbb{R}^{T_{\mathrm{dx}} \times F_{\mathrm{dx}}}$, $\mathbf{X}^{\mathrm{drug}} \in \mathbb{R}^{T_{\mathrm{dr}} \times F_{\mathrm{dr}}}$, $T_{(\cdot)}$ denotes the number of time steps and $F_{(\cdot)}$ denotes the feature dimension. 

Each temporal modality is accompanied by a global time index $g^{(\cdot)} \in [0,1]^{T_{(\cdot)}}$, which represents $T_{(\cdot)}$ irregular steps across one or multiple clinical visits within the full look-back period for each patient. Due to the on-demand nature of laboratory testing, a binary mask $\mathbf{M} \in \mathbb{R}^{T_{\mathrm{lab}} \times F_{\mathrm{lab}}}$ is also provided, where $\mathbf{M}_{i,j}=1$ indicates that the $j$-th lab test at time step $i$ is observed, and $0$ otherwise.

\noindent \textbf{Modality Encoding.} Next, we introduce the encoding component for each modality, as shown in Figure~\ref{fig:framework}. Each encoder maps the raw input $\mathbf{X}^{(\cdot)}$ into a compact latent representation. For temporal modalities, the encoder further accounts for irregular time steps and missing patterns.

\textbf{\emph{Demographics Encoder}.}
For static demographic features $\mathbf{X}^{\mathrm{demo}}$, we apply a two-layer feed-forward network to embed these attributes into a latent representation with dimensionality $d_{\mathrm{demo}}$: 
$$
\mathbf{H}^{\mathrm{demo}} = \operatorname{FFN}(\mathbf{X}^{\mathrm{demo}}) \in \mathbb{R}^{d_\mathrm{demo}}
$$ which provides static contextual information that anchors the integration with dynamic temporal signals.

\textbf{\emph{Irregular Diagnosis Encoder}.} Given the sparse and irregular nature of clinical encounters, each diagnosis sequence is modeled using a Gated Recurrent Unit (GRU)~\cite{chung2014empirical} to iteratively capture dependencies among diagnostic events. To account for non-uniform intervals within and across visits, the global time index $g^{\mathrm{dx}} \in [0,1]^{N_{\mathrm{dx}}}$ is concatenated with the input feature vector at each step, thereby informing the recurrent process:
$$
\mathbf{H}^{\mathrm{dx}} = \operatorname{GRU}(\mathbf{X}^{\mathrm{dx}}\|g^{\mathrm{dx}}) \in \mathbb{R}^{d_{\mathrm{dx}}}
$$ 
where $\|$ denotes the concatenation operator across feature dimension, $d_{\mathrm{dx}}$ denotes the dimensionality of obtained representations.

\textbf{\emph{Irregular Missing-aware Lab Test Encoder.}}
We first construct the lab-specific temporal embedding $\mathbf{P}$ as a trigonometric function of the global time index $g^{\mathrm{lab}}$, modulated by learnable frequency components, which amortizes the effect of missing values in individual irregular lab test records.
$$
\mathbf{P} = \mathbf{A}\,\boldsymbol{\Psi}_{\mathrm{sin}}^{\top}(g^{\mathrm{lab}}) + \mathbf{B}\,\boldsymbol{\Psi}_{\mathrm{cos}}^{\top}(g^{\mathrm{lab}}) 
\in \mathbb{R}^{T_{\mathrm{lab}} \times F_{\mathrm{lab}}},
$$
where $\mathbf{A}, \mathbf{B} \in \mathbb{R}^{T_{\mathrm{lab}} \times K}$ are learnable coefficient matrices, and 
$\boldsymbol{\Psi}_{\mathrm{sin}}, \boldsymbol{\Psi}_{\mathrm{cos}} \in \mathbb{R}^{K \times F_{\mathrm{lab}}}$ are Fourier basis functions applied to global time index $g^{\mathrm{lab}}$, with $K\ll T_{\mathrm{lab}}$ components.

To enrich the representations, we perform a masked dimension extension: the observed lab value at each time step is fused with the temporal embedding and lifted into a higher-dimensional space via a generic two-layer feed-forward network, while the missing values are encoded solely by the temporal embedding:
$$ 
\boldsymbol{h}^{\mathrm{lab,0}}_{i,j} =
\begin{cases}
\operatorname{FFN}\!\left(\boldsymbol{x}^{\mathrm{lab}}_{i,j}, \boldsymbol{p}_{i,j}\right), & \mathbf{M}_{i,j} = 1 \\
\operatorname{FFN}\!\left(\boldsymbol{p}_{i,j}\right), & \mathbf{M}_{i,j} = 0
\end{cases}
$$
This design enables missing lab entries to be imputed through learnable time-informed embeddings, while simultaneously augmenting observed values with temporal context. The fused representations $\mathbf{H}^{\mathrm{lab,0}}$ augmented with $g^{\mathrm{lab}}$ are then encoded by a GRU to capture the temporal evolution of lab test trajectories.
$$
\mathbf{H}^{\mathrm{lab}} = \operatorname{GRU}(\mathbf{H}^{\mathrm{lab,0}}\|g^{\mathrm{lab}}) \in \mathbb{R}^{d_{\mathrm{lab}}}
$$

\textbf{\emph{Irregular Drug Encoder.}}
Given the high-dimensional and irregular daily drug records, 
we concatenate the input with global time index to enrich temporal awareness. We apply two-dimensional convolutional neural networks whose kernels span the entire feature dimension, while sliding along the temporal axis. 
This design captures local temporal dependencies together with global co-usage patterns across medications.
Global average pooling is subsequently applied to obtain a compact representation:
$$
\mathbf{H}^{\mathrm{drug}} = \operatorname{Conv2D}({\mathbf{X}}^{\mathrm{drug}}\|g^{\mathrm{drug}}) \in \mathbb{R}^{d_{\mathrm{drug}}}.
$$
Empirically, this design outperforms GRU-based alternatives on the drug modality, as convolutional filters better exploit 
the high-dimensional co-occurrence structure while alleviating sparsity.

\noindent \textbf{Cross-modal Fusion for GVHD Prediction.} 
With modality-specific representations  $\mathbf{H}^{\mathrm{demo}}\in \mathbb{R}^{d_{\mathrm{demo}}}$, $\mathbf{H}^{\mathrm{dx}}\in \mathbb{R}^{d_{\mathrm{dx}}}$, $\mathbf{H}^{\mathrm{lab}}\in \mathbb{R}^{d_{\mathrm{lab}}}$, and $\mathbf{H}^{\mathrm{drug}}\in \mathbb{R}^{d_{\mathrm{drug}}}$, we first perform an optional per-modality dimension alignment via MLP to obtain a common embedding size $d$:
$$
\tilde{\mathbf{H}}^{m} = \Phi_m(\mathbf{H}^{m}) \in \mathbb{R}^{d}, 
\quad m\in\{\mathrm{demo,lab,dx,drug}\}
$$ 
After alignment, these representations are stacked into a unified collection of modality vectors.
$$
\mathbf{Z} =\big[\tilde{\mathbf{H}}^{\mathrm{demo}}\|\tilde{\mathbf{H}}^{\mathrm{lab}}\|
\tilde{\mathbf{H}}^{\mathrm{dx}}\|\tilde{\mathbf{H}}^{\mathrm{drug}}\big] 
\in \mathbb{R}^{4\times d}
$$

To contextualize salient cross-modal patterns for GVHD prediction, we project $\mathbf{Z}$ into queries, keys, and values, and apply a multi-head scaled dot-product attention with residual connections and normalization. This design captures cross-modal interactions~\cite{jiang2025multi}, uncovering predictive signals for GVHD from the interplay of lab trajectories, medications, and diagnostic context.
$$
\mathbf{Z}^{\prime} = \operatorname{LayerNorm}\left(\mathbf{Z} + \operatorname{MultiHeadAttention}(\mathbf{Z})\right).
$$

Finally, the fused representation $\mathbf{Z}^{\prime}$ is flattened, fed into a two-layer feed-forward neural networks and activated by a sigmoid function:
$$
\hat{y} = \operatorname{Sigmoid}\left(\mathrm{FFN}\!\left(\mathbf{Z}^{\prime}\right)\right)
$$
where $\hat{y}\in[0,1]$ denotes the prediction score of GVHD.

\noindent \textbf{Learning Objective.} 
Given the extreme imbalance of GVHD outcomes, we adopt a direct AUC optimization framework instead of conventional cross-entropy. 
Specifically, we employ the Pairwise AUC Margin Loss~\cite{yuan2021large,yuan2023libauc}, which encourages higher prediction scores for positive (GVHD) patients compared to negative ones. 
Formally, for a positive instance $i$ and negative instance $j$, the objective is:
$$
\mathcal{L}_{\mathrm{AUC}} = \frac{1}{|\mathcal{P}||\mathcal{N}|} 
\sum_{i \in \mathcal{P}} \sum_{j \in \mathcal{N}} 
 \log\!\left(1 + \exp\big(-(\hat{y}_i - \hat{y}_j)\big)\right)
$$
where $\mathcal{P}$ and $\mathcal{N}$ denote positive and negative sets, respectively,
$\exp(\cdot)$ is the exponential function $e^x$, and $\log(\cdot)$ is the natural logarithm.

\begin{table*}[t]
\centering
\caption{Performance Evaluation of the proposed method and baselines.}
\begin{tabular}{ccccc}
\toprule
 & \multicolumn{4}{c}{\textbf{Graft-versus-host disease (GVHD)}}\\
\cmidrule(lr){2-5}
\textbf{Method $\downarrow$} & \textbf{AUC} & \textbf{AUPRC} & \textbf{Recall} & \textbf{Specificity} \\
\midrule
\small \textbf{Logistic Reg.} & 0.563 $\pm$ 0.003 & 0.053 $\pm$ 0.008 & 0.358 $\pm$ 0.060 & 0.808 $\pm$ 0.056 \\
\textbf{SVC} & 0.723 $\pm$ 0.010 & 0.055 $\pm$ 0.003 & 0.719 $\pm$ 0.029 & 0.627 $\pm$ 0.022 \\
\textbf{Random Forest} & 0.633 $\pm$ 0.006 & 0.048 $\pm$ 0.008 & 0.374 $\pm$ 0.051 & 0.789 $\pm$ 0.039 \\
\textbf{AdaBoost} & 0.617 $\pm$ 0.030 & 0.039 $\pm$ 0.009 & 0.615 $\pm$ 0.149 & 0.605 $\pm$ 0.090 \\
\textbf{Gradient Boost} & 0.513 $\pm$ 0.018 & 0.031 $\pm$ 0.003 & 0.162 $\pm$ 0.128 & \textbf{0.814 $\pm$ 0.084} \\
\textbf{XGBoost} & 0.660 $\pm$ 0.013 & 0.054 $\pm$ 0.006 & 0.657 $\pm$ 0.020 & 0.646 $\pm$ 0.004 \\

\midrule
\textbf{RAIM} & 0.710 $\pm$ 0.022 & 0.135 $\pm$ 0.014 & 0.644 $\pm$ 0.066 & 0.688 $\pm$ 0.053 \\

\textbf{F-LSTM} & 0.738 $\pm$ 0.012 & 0.107 $\pm$ 0.015 & 0.676 $\pm$ 0.067 & 0.703 $\pm$ 0.060 \\
\textbf{F-CNN}  & \underline{0.764 $\pm$ 0.012} & \underline{0.147 $\pm$ 0.008} & \underline{0.737 $\pm$ 0.015} & 0.651 $\pm$ 0.011 \\
\textbf{F-Transformer} & 0.718 $\pm$ 0.009 & 0.072 $\pm$ 0.009 & 0.732 $\pm$ 0.087 & 0.606 $\pm$ 0.073 \\
\textbf{Dipole}  & 0.725 $\pm$ 0.006 & 0.120 $\pm$ 0.021 & 0.715 $\pm$ 0.040 & 0.641 $\pm$ 0.057 \\

\midrule
\textbf{Ours} & \textbf{0.836 $\pm$ 0.007} & \textbf{0.157 $\pm$ 0.014} & \textbf{0.768 $\pm$ 0.061} & \underline{0.803 $\pm$ 0.053} \\


\bottomrule
\end{tabular}%
\label{tab::main_results}
\end{table*}

To improve efficiency under severe imbalance, we use a dual sampler that adaptively pairs each positive case with a dynamically selected subset of negatives, thereby avoiding exhaustive pairwise comparisons while preserving informative gradients. Optimization is carried out with the Proximal Epoch Stochastic Gradient (PESG) optimizer~\cite{yuan2021large}, which integrates stochastic gradient descent with dual variable updates for provable convergence and stable AUC maximization.
Importantly, this AUC-oriented objective emphasizes correct ranking of high-risk GVHD patients against the majority of non-GVHD cases, aligning directly with the clinical need for early identification of rare but life-threatening complications. This prioritization ensures the model remains sensitive to early warning signals, even when positive cases are extremely scarce.
 
\subsection*{Baselines, Implementation, and Experiment Setup}
We first introduce the details for replication of multi-modal baseline methods. For traditional baselines, we construct multi-modal patient-level features by aggregating across all available modalities. Demographics are directly included as static covariates. For diagnoses, we apply a simple binary indicator, marking whether a patient has any diagnosis codes across the observation window. Laboratory results are first filled using mean imputation across the cohort. From the imputed values, we compute mean, minimum, and maximum for each lab test across time, and concatenate these as fixed-length features. For drugs, we collapse drug exposures by summing over time, yielding total counts per drug. Finally, the selected modality-specific features are concatenated into a single vector. This design reduces heterogeneous temporal EHR data into standardized, patient-level inputs that can be directly consumed by traditional machine learning models (\textit{e.g.}, logistic regression, random forest, boosting methods). We implemented deep learning baselines including RAIM~\cite{xu2018raim}, Dipole~\cite{ma2017dipole}, F-LSTM/CNN~\cite{tang2020democratizing}, and a Transformer-based variant (F-Transformer). For these baselines, we iterate over three temporal modalities as the major modality while aggregating the others as covariates (following the same procedure used in the traditional baselines) for cross-modal integration and interaction. We reported the best performance across the three configurations, using the AUC score as the primary evaluation criterion. All deep learning baselines are also optimized using Pairwise AUC Margin Loss to alleviate class imbalance.

Next, we introduce the experiment settings for evaluation. All input features are standardized using a standard scaler. The number of learnable frequency components is set to $K = 12$. All modality encoders use a hidden dimension of 32 ($d_{\mathrm{demo}} = d_{\mathrm{lab}} = d_{\mathrm{dx}} = d_{\mathrm{drug}} = 32$), and the feed-forward layers have a hidden size of 128. Training is performed with a batch size of 64 and a dual sampler ratio of 0.5 to balance positive and negative pairs, using the PESG optimizer with a learning rate of 0.001 for AUC maximization. or other deep learning baselines, we use a hidden size of 128 and optimize with Adam~\cite{kingma2014adam} using the same batch size and learning rate. We adopt stratified 5-fold cross-validation that preserves the class imbalance, repeated over 3 random seeds. We report the mean and standard deviation of AUC, AUPRC, Recall, and Specificity.




\begin{table*}
\centering
\caption{The ablation study of input modalities for early GVHD prediction.}
\begin{tabular}{cccccccc}
\toprule
\multicolumn{4}{c}{\textbf{Modalities}} & \multicolumn{4}{c}{\textbf{Graft-versus-host disease (GVHD)}}\\
\cmidrule(lr){1-4} \cmidrule(lr){5-8}
\textbf{Demo.} & \textbf{Dx.} & \textbf{Lab.
} & \textbf{Drug} & \textbf{AUC} & \textbf{AUPRC} & \textbf{Recall} & \textbf{Specificity} \\
\midrule
\cmark  & \cmark & \xmark & \xmark &  0.708 $\pm$ 0.032 & 0.081 $\pm$ 0.038 & 0.611 $\pm$ 0.079 & 0.703 $\pm$ 0.044  \\
\cmark  & \xmark & \cmark & \xmark &  0.692 $\pm$ 0.008 & 0.061 $\pm$ 0.015 & 0.680 $\pm$ 0.056 & 0.642 $\pm$ 0.066  \\
\cmark  & \xmark & \xmark & \cmark &  0.762 $\pm$ 0.009 & 0.118 $\pm$ 0.013 & 0.585 $\pm$ 0.095 & \textbf{0.817 $\pm$ 0.051} \\
\cmark  & \cmark & \cmark & \xmark &  0.754 $\pm$ 0.011 & 0.060 $\pm$ 0.007 & \underline{0.729 $\pm$ 0.036} & 0.659 $\pm$ 0.057  \\
\cmark  & \xmark & \cmark & \cmark & 0.771 $\pm$ 0.008 & 0.128 $\pm$ 0.030 & 0.681 $\pm$ 0.062 & 0.756 $\pm$ 0.041   \\
\cmark  & \cmark & \xmark & \cmark &  \underline{0.822 $\pm$ 0.017} & \underline{0.147 $\pm$ 0.012} & 0.671 $\pm$ 0.038 & \underline{0.813 $\pm$ 0.064}  \\
\midrule
\cmark  & \cmark & \cmark & \cmark & \textbf{0.836 $\pm$ 0.007} & \textbf{0.157 $\pm$ 0.014} & \textbf{0.768 $\pm$ 0.061} & 0.803 $\pm$ 0.053   \\ 
\bottomrule
\end{tabular}%
\label{tab::modality_ablation}
\end{table*}

\section*{Results}

\subsection*{Results of Baseline Comparisons, Ablation Study, and Model Analysis}
Table~\ref{tab::data_summary} summarizes the characteristics of the four electronic health record modalities included in this study. Demographic data consisted of 4 static features. Laboratory data consisted of 74 features measured across 24 one-hour intervals. Diagnosis data comprised 20 features aggregated into 12 steps, each representing a 3-month interval. Medication (Drug) data included 280 features recorded over 24 steps with a daily resolution. The granularity and temporal ranges were determined in consultation with clinical experts and validated through empirical experiments for each modality.

Table~\ref{tab::main_results} presents the performance of the proposed framework compared with traditional machine learning methods and deep learning baselines for early GVHD prediction. Among the traditional machine learning methods, Support Vector Classifier (SVC) achieved the highest AUC (0.723 $\pm$ 0.010) and recall (0.719 $\pm$ 0.029), while Gradient Boosting obtained the highest specificity (0.814 $\pm$ 0.084). Overall, deep learning baselines generally outperformed traditional methods, particularly in terms of AUC and AUPRC. Within the deep learning group, F-CNN yield the stronger performance in terms of AUC (0.764 $\pm$ 0.012) and recall (0.737 $\pm$ 0.015), whereas TANet reached the highest AUPRC (0.148 $\pm$ 0.016). Notably, the proposed framework demonstrated the most consistent improvements across evaluation metrics, achieving the best overall AUC (0.836 $\pm$ 0.007), the highest AUPRC (0.157 $\pm$ 0.014), a leading recall (0.768 $\pm$ 0.061), and the second-best specificity (0.803 $\pm$ 0.053).

Table~\ref{tab::modality_ablation} reports the ablation results evaluating the contribution of different modalities for early GVHD prediction. Using demographics combined with diagnoses achieved an AUC of 0.708 $\pm$ 0.032, while demographics with laboratory tests reached an AUC of 0.692 $\pm$ 0.008. Incorporating drugs along with demographics resulted in stronger performance (AUC 0.762 $\pm$ 0.009, AUPRC 0.118 $\pm$ 0.013), with the highest specificity observed in this setting (0.817 $\pm$ 0.051). Overall, adding multiple modalities consistently enhanced performance. For instance, combining demographics, diagnoses, and drugs yielded an AUC of 0.822 $\pm$ 0.017 and an AUPRC of 0.147 $\pm$ 0.012, both higher than any two-modality configuration. The best overall performance was observed when all four modalities were integrated, with results surpassing all ablation settings.

\begin{figure*}
    \centering
    \includegraphics[width=0.92\linewidth]{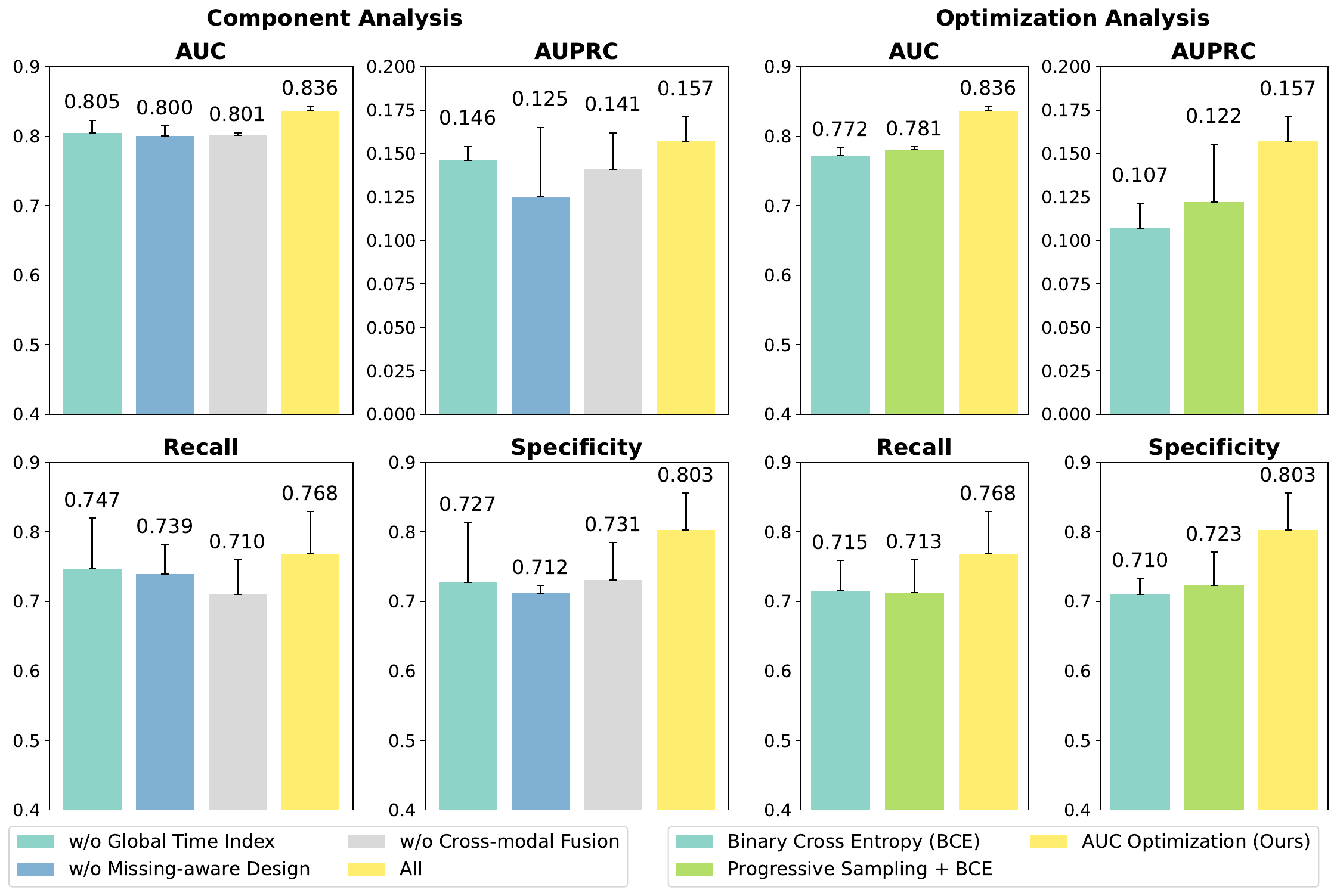}
    \caption{Left: Analysis of model components, Right: Comparison of optimization methods handling class imbalance.}
    \label{fig:comb}
\end{figure*}

The left subplot of Figure~\ref{fig:comb} illustrates the contribution of different model components to early GVHD prediction. Removing the global time index resulted in a decrease of AUC to 0.805 and AUPRC to 0.146, with recall and specificity at 0.747 and 0.727, respectively. Excluding the missing-aware design further reduced AUC to 0.800 and AUPRC to 0.125, accompanied by recall of 0.739 and specificity of 0.712. Without the cross-modal fusion module, the model achieved an AUC of 0.801 and an AUPRC of 0.141, with recall of 0.710 and specificity of 0.731. When all components were included, our model achieved the strongest performance.

The right subplot compares different optimization strategies for handling class imbalance in early GVHD prediction. Using Binary Cross Entropy (BCE) alone resulted in an AUC of 0.772, an AUPRC of 0.107, a recall of 0.715, and a specificity of 0.710. Incorporating progressive sampling with BCE improved performance slightly, yielding an AUC of 0.781, an AUPRC of 0.122, recall of 0.713, and specificity of 0.723. The AUC optimization consistently achieved the best performance across all evaluation metrics, improving by a clear margin compared with the other variants. 

\section*{Discussion}

In this study, we developed a multi-modal deep learning framework to predict GVHD prior to liver transplantation surgery, using the EHRs of a real-world cohort from Mayo Clinic. Based on the reported results, we highlighted three key observations: (1) \textbf{Complementary predictive value of modalities.} Different temporal modalities provide distinct but complementary signals for early GVHD prediction, where effective integration strategies are crucial. (2) \textbf{Importance of handling data irregularity and missingness.} Robust methods that account for irregular and missing records improve predictive performance. (3) \textbf{Addressing severe class imbalance.} The extreme imbalance between GVHD and non-GVHD cases hinders model performance. Our integration of direct AUC optimization proved effective in mitigating this issue, leading to more reliable early detection.

Our first key finding is that multi-modal EHR data capture diverse clinical signals essential for early GVHD prediction. As shown in Table~\ref{tab::modality_ablation}, incorporating additional modalities consistently improves predictive performance across all combinations. Among the individual temporal sources, drug exposure data stood out with a relatively high AUC (0.762 $\pm$ 0.009) and the highest specificity (0.817 $\pm$ 0.051), suggesting that medication patterns prior to transplant may contain strong predictive cues for downstream complications. Diagnosis history also contributed positively, especially when combined with drug data, yielding the second-best overall performance (AUC: 0.822 $\pm$ 0.017; AUPRC: 0.147 $\pm$ 0.012; Specificity: 0.813 $\pm$ 0.064). This synergy likely reflects the clinical context in which diagnosis and medication co-evolve, providing a proxy for evolving patient status. Adding laboratory measurements further strengthened predictive accuracy, as they capture dynamic indicators of immune activation, organ dysfunction, and other early clinical signs of GVHD, culminating in the best results when all four modalities are used.

As reported in Table~\ref{tab::main_results}, our method outperformed both traditional machine learning and deep learning baselines. Traditional models, which rely on summarizing statistics concatenated across modalities, are limited in their ability to capture temporal dependencies. Similarly, several deep learning baselines focus on one primary temporal stream while incorporating other modalities as static covariates for cross-modal interactions. In contrast, our framework performs end-to-end learning across all temporal modalities, enabling joint modeling of fine-grained temporal patterns and inter-modality dependencies. This design enables richer cross-modal interactions, such as associating changes in diagnosis with concurrent medication adjustments, thereby uncovering subtle yet clinically meaningful patterns. By using self-attention across modalities, our framework learns to align temporally heterogeneous signals in a dynamic manner. As shown in Figure~\ref{fig:comb}, removing the cross-modal fusion mechanism resulted in a notable drop in AUC (from 0.836 to 0.801) and AUPRC (from 0.157 to 0.141), underscoring the importance of effective integration in capturing synergistic signals across heterogeneous clinical modalities.

Our second key observation underscores the importance of robust mechanisms for handling data irregularity and missingness in multi-modal EHRs. As shown in the component analysis of Figure~\ref{fig:comb}, removing either the global time index or the missing-aware design resulted in consistent performance degradation across all evaluation metrics. The global time index serves as a unified temporal reference that aligns asynchronous events across modalities, while the missing-aware design captures patterns of informative absence, which may reflect underlying clinical intent or patient stability. Notably, removing the missing-aware component led to a substantial drop in AUPRC, from 0.157 to 0.125, reflecting degraded detection of rare GVHD cases under imbalance.

Our third key observation emphasizes the importance of directly addressing the extreme class imbalance in GVHD prediction.
As shown in the optimization analysis of Figure~\ref{fig:comb}, conventional training using BCE loss yields suboptimal performance, particularly in AUPRC (0.107). Even when combined with progressive sampling  that emphasizes rare positive instances in the early training stage, the improvement is marginal. In contrast, our approach using direct AUC optimization achieves a substantial boost in both AUC (from 0.772 to 0.836) and AUPRC (from 0.107 to 0.157), indicating better ranking and detection of rare positive cases. Additionally, specificity improves from 0.710 to 0.803 without sacrificing recall. The reported performance gains underscore the value of AUC optimization in learning more discriminative decision boundaries for early GVHD prediction.

\section*{Conclusion Remark and Future Work}

In this study, we present a multi-modal deep learning framework for early prediction of graft-versus-host disease (GVHD) prior to liver transplantation, using heterogeneous pre-transplant EHR data. Using a real-world cohort from Mayo Clinic, we integrated demographics, laboratory tests, diagnoses, and medications into a unified predictive model. Through modality-specific encoding, cross-modal fusion, and direct AUC optimization, our method achieved strong performance under extreme class imbalance. Extensive evaluations underscore the advantages of our proposed method over baselines and the effectiveness of its components, showing the potential of early GVHD detection in liver transplantation.

This work has two primary limitations: (1) potential cohort bias, as the dataset was derived from a single institution, and (2) limited model interpretability. In future work, we plan to expand the cohort by incorporating more diverse patient populations from the Mayo Clinic Platform~\cite{MayoClinicPlatform2025} and to enhance explainability through case-based reasoning~\cite{jiang2025timexl,ni2021interpreting,jiang2023interpretable} and modality gating mechanisms~\cite{jiang2022cross,yun2024flex} that quantify feature contributions across data modalities.

\section*{Funding}
This project was funded by the National Institute of Health, grant number R01AG083039. It was performed during the first author's summer research internship at Mayo Clinic AI \& I department. Dongjin Song gratefully acknowledges the support of the National Science Foundation under Grant No. 2338878, as well as the generous research gifts from NEC Labs America and Morgan Stanley.

\bibliographystyle{vancouver}
{\footnotesize \bibliography{citation}}

\end{document}